\documentclass[10pt,twocolumn,letterpaper]{article}

\usepackage{iccv}
\usepackage{times}
\usepackage{epsfig}
\usepackage{graphicx}
\usepackage{amsmath}
\usepackage{amssymb}
\usepackage{multirow}
\usepackage{booktabs}
\usepackage{pifont}
\usepackage{subcaption}
\usepackage{dashrule, arydshln}
\usepackage{colortbl} 
\usepackage{enumitem}

\newcommand{\xmark}{\ding{55}}

\newcommand{\metricName}{{Entropic Score}}
\newcommand{\metricNameSpace}{{Entropic Score} }

\definecolor{Gray}{gray}{0.9}

\usepackage[pagebackref=true,breaklinks=true,colorlinks,bookmarks=false]{hyperref}

\iccvfinalcopy 

\def\iccvPaperID{53} 

\ificcvfinal\fi

\begin{document}

\title{\vspace{-1cm} Entropic Score metric: Decoupling Topology and Size in Training-free NAS}

\author{
Niccol\`{o} Cavagnero\\
\and
Luca Robbiano\\
\and
Francesca Pistilli\\
\and
Barbara Caputo\\
\and
Giuseppe Averta\\
{\tt\small name.surname@polito.it}\\
Politecnico di Torino, Corso Duca degli Abruzzi, 24 - 10129 Torino, ITALIA\\
}

\maketitle
\ificcvfinal\thispagestyle{empty}\fi

\begin{abstract}
Neural Networks design is a complex and often daunting task, particularly for resource-constrained scenarios typical of mobile-sized models. Neural Architecture Search is a promising approach to automate this process, but existing competitive methods require large training time and computational resources to generate accurate models. 
To overcome these limits, this paper contributes with: i) a novel training-free metric, named \metricName, to estimate model expressivity through the aggregated element-wise entropy of its activations; ii) a cyclic search algorithm to separately yet synergistically search model size and topology.
\metricNameSpace shows remarkable ability in searching for the topology of the network, and a proper combination with LogSynflow, to search for model size, yields superior capability to completely design high-performance Hybrid Transformers for edge applications in less than 1 GPU hour, resulting in the fastest and most accurate NAS method for ImageNet classification. Code available here\footnote{\hyperlink{https://github.com/NiccoloCavagnero/EntropicScore}{https://github.com/NiccoloCavagnero/EntropicScore}}.
\end{abstract}

\section{Introduction}
\label{intro}
\begin{figure}[t]
    \centering
        \includegraphics[width=\columnwidth]{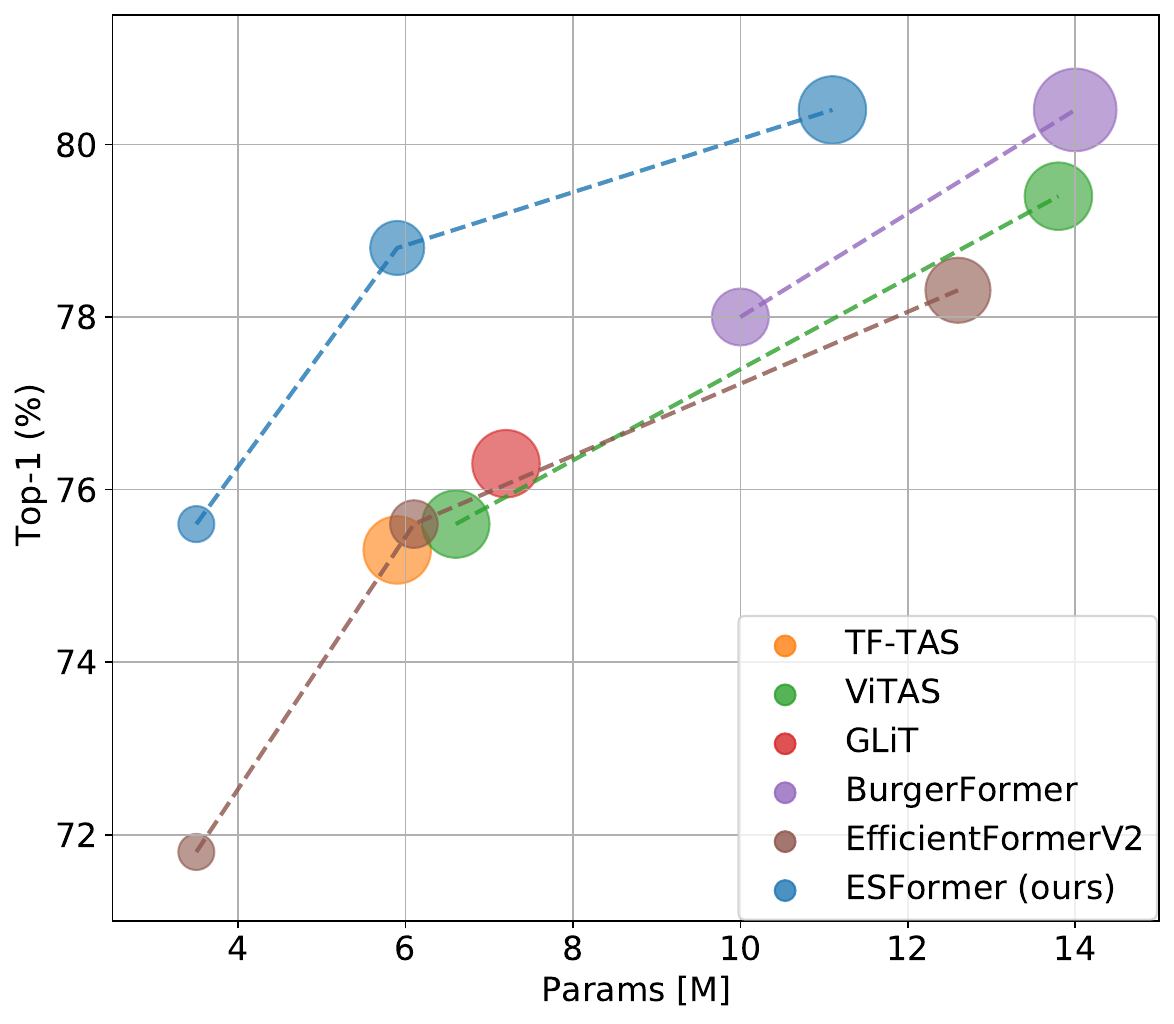}
    \caption{Model Size vs ImageNet-1k Top-1 Accuracy for state-of-the-art NAS methods. The size of each point represents the model's MACs.}
    \label{fig:teaser}
\end{figure}
The design of neural networks has been a pivotal research area in deep learning, with many notable examples~\cite{resnet, resnext, mobilenetV2, efficientnet, convnext, vit}. In an attempt to foster deep learning on edge applications, in the last few years there has been a particular interest of the community for the development of tiny architectures able to efficiently run on limited-resource hardware, such as mobile devices.

However, the manual design of such models is a challenging task, further exacerbated by the need of finding a trade-off between model accuracy and computational efficiency. This is especially true for Transformer-based architectures~\cite{attention, vit}, which suffer from quadratic increase in computational complexity as the size of input data grows. As a result, deploying such models in resource-constrained environments can be extremely challenging.

Neural Architecture Search (NAS) has emerged as an effective tool to automate this process at the expenses of long and costly training phases to evaluate all the candidate networks~\cite{nasvit, glit, vitas, efficientFormerV2}, which make the search process computationally expensive and time-consuming.

Recently, training-free approaches~\cite{naswot, tenas, freerea, zenscore, dss} have been proposed to simplify and speed-up the neural architecture search process.
The core idea is to completely replace the training phase with the computation of zero-shot metrics to score the networks at initialisation. 

Although these solutions offer a significant reduction in computation time and cost, most of the metrics proposed so far only encode specific characteristics of the network, and their adoption for the design of the whole architecture potentially leads to sub-optimal models.

To push the boundaries of training-free NAS, then, it is crucial to provide better metrics, more strictly related to relevant model attributes, such as its dimensionality and topology, that can be adopted for improved decoupled search strategies, where each metric only drives the design of specific model characteristics. 

In this paper, we propose a solution to these problems with a novel training-free NAS algorithm where an original metric, \metricName, is introduced to co-supervise the search process.
\metricNameSpace captures the expressivity of the candidate models by means of an entropy-like function over the activation layers' outputs, showing to be particularly suited for the design of the network topology, an aspect of paramount importance for the accuracy of the searched architecture.\par

Furthermore, the search algorithm relies on a novel decoupled design paradigm, which synergistically yet independently designs model size and topology based on a proper combination of \metricNameSpace and LogSynflow~\cite{freerea}, an advanced variant of the popular Synflow~\cite{synflow}.

Unlike previous approaches~\cite{tenas, freerea, zerocost, dss}, which seamlessly combine metrics in an aggregated score, we propose to decouple two aspects of the network design, the topology and the dimensionality, supervising each search with a dedicated metric, \metricNameSpace and LogSynflow~\cite{freerea} respectively. This strategy enables a more targeted search and a better exploitation of the strengths of each metric.

The experimental results demonstrate the effectiveness of our approach in discovering high-performing neural networks without the need for training, improving the accuracy and the efficiency of the search. 
The resulting models perform favourably not only with respect to hand-designed architectures but also with respect to training-based NAS methods (see Figure~\ref{fig:teaser}). 

Remarkably, the search process requires less than 1 GPU hour, highlighting the efficiency of our training-free algorithm and enabling the design of resource-efficient Hybrid Transformers in a timely manner.

To conclude, this paper contributes with: 
\begin{itemize}[noitemsep]
    \item a new data-agnostic metric, named \metricName, for the assessment of model topology;
    \item a decoupled search strategy to fully exploit the potential of two complementary metrics for neural networks design, capable to accurately tailor model dimension and topology in less than 1 GPU hour;
    \item a thorough experimental validation, together with the release of ESFormers, a family of tiny Hybrid Transformers that outperforms existing mobile-sized models for ImageNet classification.
\end{itemize}
\section{Related Works}
\label{related}

\subsection{Hybrid Transformers}
The advent of Transformers~\cite{attention} marked a significant milestone in deep learning, where the multi-head attention mechanism has been successfully applied to various domains obtaining state-of-the-art results. 

Nevertheless, when it comes to Computer Vision tasks, Vision Transformers (ViTs)~\cite{vit} lack some of the critical inductive biases present in Convolutional Neural Networks (CNNs), such as translation equivariance and locality. This crucial drawback leads to a need for significantly larger amount of data~\cite{vit} or longer and more sophisticated training pipelines~\cite{deit} to match similar performances. 

Furthermore, the classic attention mechanism does not enjoy weight sharing and it scales quadratically with respect to the input dimension. This poses critical difficulties in adopting Transformer-based architectures in mobile settings or downstream tasks that require large input signals.

To address these challenges, the research community has been focusing on two main research fields: developing more efficient attention mechanisms or combining Convolutional Neural Networks and Transformers to exploit the strengths of both architectures.

One noticeable example of the first approach is Swin Transformer~\cite{swin}, which employs a local window to improve efficiency at the expense of the global receptive field of standard attention. Other following studies~\cite{xcit,mobilevit,efficientFormerV2} have proposed alternative attention mechanisms that can be used to improve the speed and performance trade-off of Transformer-based architectures.

Instead, the hybridisation of CNNs and Transformers aims to directly incorporate convolutional biases into the Transformer architecture by combining convolutions and attention in a single model.

CoAtNet~\cite{coatnet} is a pioneering example of a CNN-Transformer hybrid, adopting Inverted Bottleneck blocks (IBN)~\cite{mobilenetV2} for the first two stages of the architecture and Transformer blocks in the last two. The resulting family of hybrid models has achieved state-of-the-art performance by outperforming both CNNs and pure ViT architectures. LocalViT~\cite{localvit} took a step forward alternating global and local computations across all Transformer blocks. Specifically, it introduces locality replacing all the standard Multi-Layer Perceptrons (MLPs) with IBNs. Other Transformer hybrids~\cite{xcit,mobilevit,edgevit,glit,efficientformer,efficientFormerV2} apply similar concepts.

Still, adopting Vision Transformers in resource-constrained scenarios remains a challenging task and different NAS approaches have been proposed to tackle this issue~\cite{dss,glit,vitas,burger,efficientformer,efficientFormerV2}.

\subsection{Neural Architecture Search}
The field of Neural Architecture Search was first introduced in a notable study~\cite{reinforcement_nas}, which employed Reinforcement Learning (RL) to generate high-performing neural networks. However, this approach requires over 22,400 GPU-hours for the partial training of tens of thousands of networks, making it prohibitively expensive from a computational point of view. Consequently, researchers have been exploring more efficient NAS methods, such as differentiable and evolution-based search techniques.

Differentiable methods aim to make the entire search process differentiable to enable optimisation using gradient descent algorithms~\cite{darts,gdas}. These approaches have led to significant improvements compared to the original RL-based method~\cite{reinforcement_nas} in terms of efficiency. 

It is worth noting that, since these methods require the use of a supernet, the dimensionality of the search space may be strongly limited due to memory constraints. Furthermore, it is not straightforward to apply differentiable methods to ViT architectures due to the presence of gradient conflicts in the supernet~\cite{nasvit}. 

On the other hand, evolution-based techniques, such as those discussed in~\cite{evo_nas_review} and~\cite{evo_large}, are easier to implement with respect to the former categories, and enable natural parameter inheritance from parent networks. However, they have been found to be less effective than other search techniques~\cite{rea}. REA algorithm~\cite{rea} introduced a regularised Tournament Selection approach, resulting in the first evolution-based NAS method able to outperform human-designed neural networks. 

Nevertheless, all these classic NAS techniques still require expensive training phases of thousands of candidate architectures. This highlights the ongoing challenges in NAS research in terms of computational efficiency, partially solved by the adoption of training-free techniques.

\subsection{Training-free NAS}
In recent years, there has been an increasing interest in training-free methods, which are known for their efficiency and scalability. A key role in this framework is played by the chosen metrics that supervise the search process acting as a proxy for the accuracy of an untrained network. To this end, several metrics have been proposed, each with its own advantages and drawbacks.

The first proposed metric was NASWOT~\cite{naswot}, a proxy for the expressivity of a network, which measures the similarity of activation patterns for different input samples. TE-NAS~\cite{tenas} improved NASWOT by incorporating the trainability of the architectures through the use of the Neural Tangent Kernel (NTK)~\cite{ntk}. However, NTK is computationally expensive, time-consuming, and it has been shown to have low correlation with accuracy~\cite{freerea,zerocost}.

The study of Zero-cost Proxies~\cite{zerocost} analysed various saliency-based metrics from pruning literature and found Synflow~\cite{synflow} to be superior with respect to other approaches~\cite{naswot,grasp,snip,fisher}. FreeREA~\cite{freerea} further enhanced Synflow by proposing LogSynflow, which adopts a logarithmic function to scale down the gradients to mitigate the issue of gradient explosion. Moreover, the authors demonstrated that the contribution of NASWOT when combined with Synflow and its variants is extremely limited.

In addition, there are two other metrics worth mentioning: Zen-score~\cite{zenscore} and DSS~\cite{dss}. Both of these metrics are correlated with the expressivity of networks. Zen-Score measures the expected Gaussian complexity of a given convolutional network, while DSS is a Synflow variant that takes into account the synaptic diversity of attention weight matrices. Nonetheless, Zen-score is specifically designed for Convolutional Neural Networks and DSS for pure Transformers architectures~\cite{dss}, and therefore they are not seamlessly adaptable for the purpose of our work.

\section{Method}
\label{method}

\subsection{Search for topology and size}

Model design can be categorised in two main families: topological and dimensional. Topology refers to the structure of the network, including the types of layers, their connections, and how they are arranged (see Figure~\ref{fig:blocks}). Size, on the other hand, refers to the number of parameters or the computational cost of the model. The latter can be controlled by the varying, for example, the number of layers, the number of channels in each layer, the expansion ratios in bottlenecks, and so on.

Following this categorisation, different NAS benchmarks have been introduced. In particular, NATS-Bench~\cite{nats} contains a topological search space of more than 15 thousands convolutional topologies and a size search space with more than 35 thousand networks with same structure and different dimensionality. NAS-Bench-101~\cite{bench-101} instead contains over 400 thousands convolutional architectures with varying topologies.

\begin{figure*}[t]
     \centering
     \captionsetup[subfigure]{justification=centering}
     \begin{subfigure}[b]{0.9\columnwidth}
         \centering
         \includegraphics[width=\columnwidth]{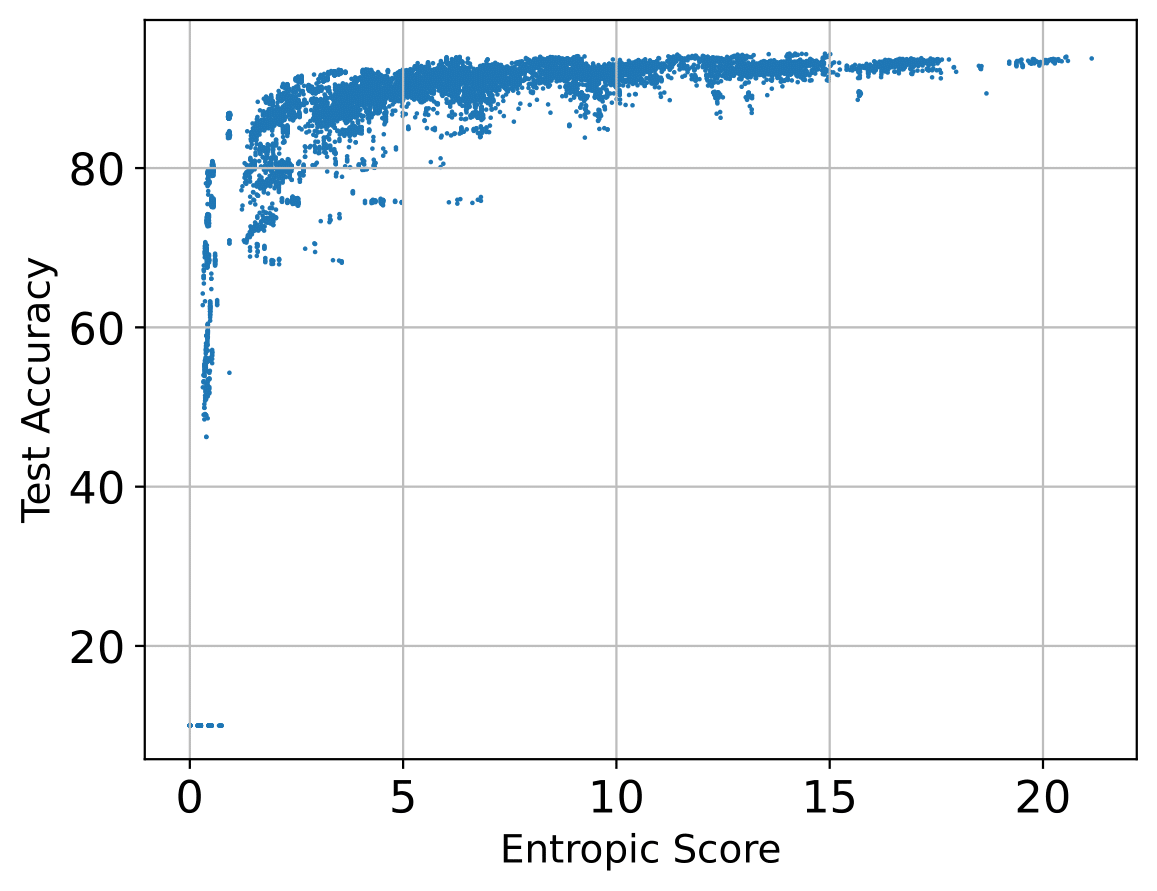}
         \caption{\textit{Topological} search space. \\   
         \metricNameSpace Spearman $\rho=0.68$.}
         \label{fig:entropy}
     \end{subfigure}
     \hfill
     \begin{subfigure}[b]{0.9\columnwidth}
         \centering
         \includegraphics[width=\columnwidth]{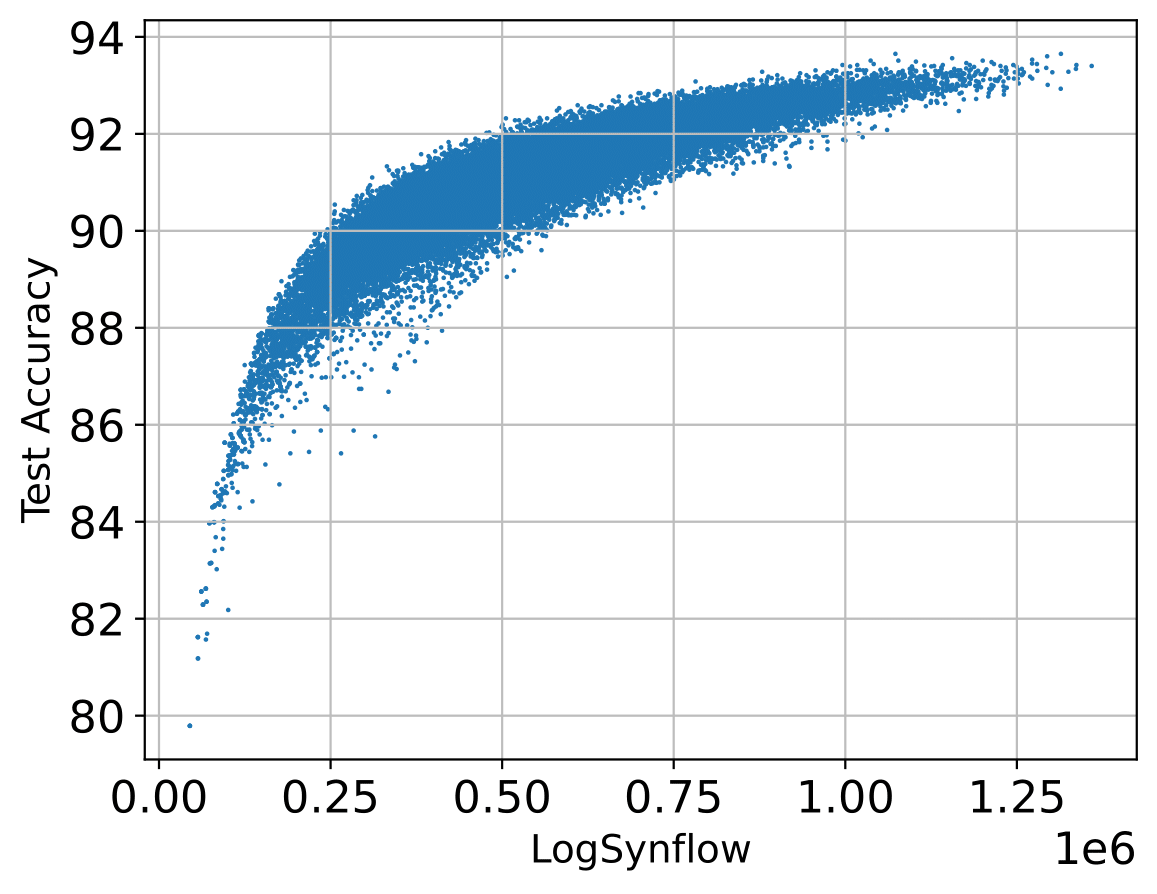}
         \caption{\textit{Size} search space. \\
         LogSynflow Spearman $\rho=0.92$.}
         \label{fig:logsynflow}
     \end{subfigure}
     \caption{Training-free metrics vs CIFAR10 test accuracy. a) \metricNameSpace evaluated on a topological search space (NAS-Bench-101~\cite{bench-101}). b) LogSynflow evaluated on a dimensional search space (NATS-Bench~\cite{nats}). Entropic Score shows to be particularly suitable in choosing topologies, while LogSynflow excels in dimensioning the architectures.}
     \label{fig:correlations}
\end{figure*}

\subsection{\metricName}

The training-free NAS method proposed in this paper exploits a novel metric, called \metricName, to guide the search process. \metricNameSpace represents a measure of the network ability to represent and encode meaningful signal information, computed by feeding a random tensor to the networks and summing the average element-wise entropy of the normalised activations. 

Intuitively, we expect that the higher is the \metricName, the larger is the information flow in the forward pass, with a positive impact on the fitting capability of a given architecture. From this standpoint, \metricNameSpace may be interpreted as a proxy for the expressivity of a network. 

Similarly to Synflow~\cite{synflow} and its variants, \metricNameSpace is completely data-agnostic. Searched architectures are therefore generic and not specifically related to a given dataset, naturally enabling the adoption of the models in different scenarios. 
Therefore, we propose a general search algorithm, not dataset-constrained, able to provide models for a given task that can be adopted in various settings. 

Given a network parameterised by $\theta$, the proposed aggregated metric can be defined as follows:
\begin{equation}
\label{score}
    E(\theta) = - \sum_{i=1}^N \frac{1}{K} \sum_{j=1}^K a_{ij}(\theta) \cdot \log{(a_{ij}(\theta) +\epsilon)},
\end{equation}

where $N$ is the number of activations in the network, $K$ is the number of elements in the normalised activation tensor $a$ and $\epsilon$ is a small stability constant. 

A critical step of the proposed \metricNameSpace consists in the computation of the layer-wise value that is later aggregated to rank the whole architecture.
Before computing the score, the network must undergo a preparation step inspired by Synflow~\cite{synflow}.
Namely, we suppress all normalisation operators and take the absolute value of the weights. Then, any activation, such as GELU~\cite{gelu} or Swish~\cite{swish}, is replaced by a ReLU function~\cite{relu}. This way, only non-negative values are propagated through the network.
Next, a random tensor $x \in [-0.5, 0.5]$ is fed to the network and the activation values are normalised in the interval $(0,1]$ by dividing for their maximum value across the channel dimension. 

The layer-wise score is computed by taking the average element-wise entropy of these normalised activation values. Finally, we aggregate the score across the layers to provide a measure for the expressivity of a network topology. In practice, we compute \metricNameSpace three times with different network and input initialisations and take the average as the actual score.

In the context of NAS, \metricNameSpace provides information about the potential expressivity of a network, as models with higher \metricNameSpace values are expected to have more complex activation patterns. \metricNameSpace proves to be particularly well-suited for designing the topology of the network (see Table~\ref{tab:correlations_topology} and Figure~\ref{fig:entropy}).

\subsection{Decoupled Search}

Since metrics for training-free NAS provide cues on different characteristics of neural models, in our method we developed a strategy to properly combine our \metricNameSpace with LogSynflow metric~\cite{freerea}, to drive the topology and size search respectively. 

LogSynflow, which proved to be sensible for dimensionality design (see Table~\ref{tab:correlations_size} and Figure~\ref{fig:logsynflow}), constitutes an improved version of Synflow~\cite{synflow}, a saliency metric derived from pruning literature, which provides information about the gradient flow and the complexity of the network. Moreover, its strong ability in dimensioning the networks provides complementary information to \metricName. 

By aggregating \metricNameSpace and LogSynflow metrics, our approach provides an original comprehensive evaluation of candidate architectures in terms of topology and size.

Seamlessly combining different metrics, as done in previous works~\cite{tenas, zerocost, freerea, dss}, could yield to sub-optimal results as these may conflict with each other. For example, a metric with a high capability in dimensioning the model can contribute poorly in topological decisions, and vice versa. 

To better exploit the strengths of each metric, we adopt a novel decoupled approach, where \metricNameSpace and LogSynflow are used separately yet synergistically to select only specific aspects of the network (see Table~\ref{tab:search_space}). 

In particular, \metricNameSpace is adopted to choose the topological characteristics of the network, such as type of block or kernel size, while LogSynflow focuses on the size of layers selecting, among other aspects, output channel dimension and expansion ratio for bottlenecks and MLPs. 

This allows each metric to have a larger influence in areas where it excels, leading to more accurate and efficient search results. Furthermore, the proposed decoupled approach provides a flexible framework that can be easily adapted to incorporate additional metrics as needed.

\subsection{Search Space}
\label{search_space}
\begin{figure}[t]
    \centering
        \includegraphics[width=\columnwidth]{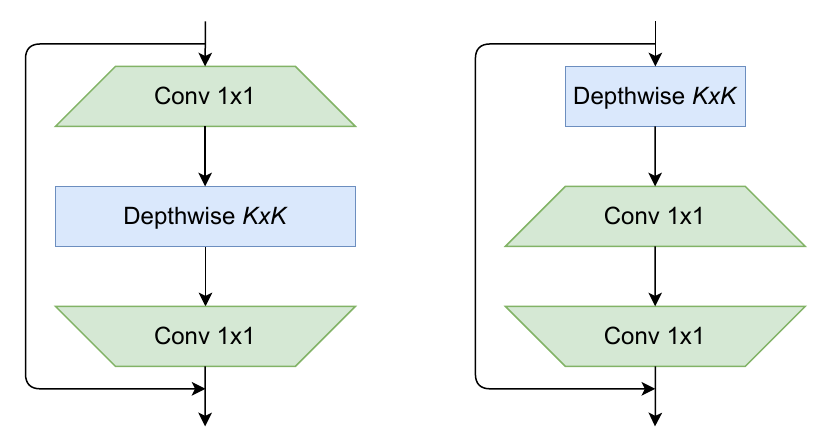}
    \caption{Different configurations of FFN blocks enhanced with locality. Left: Inverted Bottleneck Block~\cite{mobilenetV2}. Right: ConvNeXt Block~\cite{convnext}.}
    \label{fig:blocks}
\end{figure}
The search space used in this study is largely based on the design of EfficientFormerV2~\cite{efficientFormerV2}. This work defines a modern hybrid Transformer architecture that integrates Inverted Bottlenecks~\cite{mobilenetV2} and efficient Multi-head Self-Attention (MHSA) layers~\cite{efficientFormerV2}. This efficient formulation of MHSA is enhanced with downsampling, locality~\cite{nasvit, inception} and Talking Heads~\cite{talking}.

The adoption of training-free metrics allows for more effective resource utilisation, and a more detailed exploration of the network's various components is therefore feasible. To this end, the search space has been refined an expanded to allow for greater flexibility in the network design.

Specifically, the output dimension and kernel size of each Feed Forward Network (FFN) block can be independently selected, rather than having a fixed per-stage output dimension and a 3x3 kernel for the whole architecture as in~\cite{efficientFormerV2}. Moreover, the FFN can be configured not only as an IBN~\cite{mobilenetV2} but also as a ConvNeXt block~\cite{convnext} by rearranging the placement of the depthwise convolution (see Figure~\ref{fig:blocks}). 

Additionally, we also allow more flexible MHSA blocks, searching also for the number of heads and head dimension of each layer. Kernel size of FFNs following attention is instead fixed to 3x3, as there is no need for large kernel sizes given the global receptive field of MHSA.

In summary, the following elements of each FFN block are searchable: FFN configuration (IBN vs ConvNeXt), output size, kernel size, and expansion ratio. Instead, for each Transformer layer we search for FFN configuration, output size, expansion ratio, number of heads and head dimension. A detailed division between topological and size dimensions is shown in Table~\ref{tab:search_space}.

It must be noted that only non-decreasing output dimensions are allowed, such that skip connections can always be implemented by means of Zero-padded Residuals~\cite{pyramid}.

We follow~\cite{efficientFormerV2} for both the number of blocks per stage and the design of downsampling blocks.

Overall, the increased number of searchable characteristics leads to a more fine-grained and significantly larger search space with respect to the one originally proposed in EfficientFormerV2~\cite{efficientFormerV2}.
\begin{table}[t]
    \centering
    \begin{tabular}{@{}p{3.2cm}p{3.2cm}@{}}

        \toprule
        \multicolumn{1}{c}{Topology}  & \multicolumn{1}{c}{Size}  \\
        
        \midrule
        \multicolumn{1}{c}{FFN Type}            & \multicolumn{1}{c}{Output Channels}    \\
        \multicolumn{1}{c}{Kernel Size}         & \multicolumn{1}{c}{Expansion Ratio}     \\
        \multicolumn{1}{c}{Number of Heads}     & \multicolumn{1}{c}{Head Dimension}       \\
        \bottomrule
    \end{tabular}
    \caption{Division of searchable dimensions between topology and size.}
    \label{tab:search_space}
\end{table}

\subsection{Search Algorithm}
As search algorithm, we adopt an evolutionary approach based on REA~\cite{rea}. This method was further refined and improved in line with the findings of~\cite{freerea}, which introduced multiple mutations at each step and crossover operations between parent networks to escape locality and improve exploration and population diversity. All considered mutation and crossover probabilities are uniform.

Given the high dimensionality of the search space, a multi-start strategy is employed prior to the main search phase. This involves independently evolving multiple random subpopulations for a limited time, and then using the top performing architectures from each subpopulation as seeds for the main search phase. This intuitively reduces the dependence on the initial population, leading to improved results and reduced variance between different runs. 

During the initial multi-start phase, we adopt a population and a Tournament size of 25 and 5 respectively, following~\cite{freerea}. These sizes are doubled during the main search.

In order to favour better topologies from the beginning, the multi-start phase is guided solely by \metricName. The main search process then alternates between topology search and size search in a cyclic manner, determining different characteristics of the networks in each phase. The best performing models of each phase are adopted as seeds for the subsequent step. 

Specifically, the multi-start phase involves the evolution of five separate populations for 3 minutes each. The topology and size searches are then alternated every 5 minutes for the smaller models and 6 minutes for the large one, for a total search duration of 45 and 55 minutes respectively. Notably, the whole process takes less than 1 GPU hour.

\section{Experiments}
\label{experiments}

\subsection{Training Details}
In all our experiments, we evaluate the performance of the networks on the ImageNet-1k dataset~\cite{imagenet} using a standard resolution of 224x224. The networks are trained for 300 epochs with Binary Cross-Entropy using the AdamW optimiser~\cite{adamw} with a learning rate of $2e^{-3}$, a batch size of 1024, and a weight decay of $2e^{-2}$. The learning rate follows a cosine schedule~\cite{sgdr}. In order to improve stability during training, a warmup period of 20 epochs is prepended to the main training phase.

In addition, we adopt common data augmentations and regularisations, including Mixup/Cutmix~\cite{mixup, cutmix}, RandAugment~\cite{randaugment}, Random Erasing~\cite{rand_erasing}, Stochastic Depth~\cite{stochdepth}, Gradient Clipping~\cite{grad_clip} and Label Smoothing~\cite{labelsmooth}. Stronger augmentations  and regularisations are exploited to train the largest models. Detailed training hyper-parameters can be found in Table~\ref{tab:train_details}.

\begin{table}[h]
\setlength{\tabcolsep}{5.5pt}
\centering
\begin{tabular}{lcc}

\toprule
                  & S0-S1    & S2                  \\
\midrule
Optimiser         & AdamW    &   AdamW             \\
Batch Size        & 1024     &   1024              \\
LR                & 0.002    &   0.002             \\
LR schedule       & Cosine   &   Cosine            \\
Training Epochs   & 300      &   300               \\
Warmup Epochs     & 20       &   20                \\
Weight Decay      & 0.02     &   0.02              \\
Gradient Clipping & 2.0      &   0.01              \\
Mixup/Cutmix      & 0.8/1.0  &   0.8/1.0           \\
RandAugment       & m7-n1    &   m9-n1             \\
Random Erasing    & 0.0      &   0.25              \\
Stochastic Depth  & 0.05     &   0.1               \\
Label Smoothing   & 0.1      &   0.1               \\
\bottomrule
\end{tabular}
\caption{Training hyper-parameters for ImageNet-1k.}
\label{tab:train_details}
\end{table}

We do not employ distillation for faster training and for a fair comparison with previous approaches.

\subsection{ImageNet Classification}

\begin{table*}[t]
\setlength{\tabcolsep}{3.5pt}

\centering
\begin{tabular}{@{}lccccccc@{}}
\toprule
\multicolumn{1}{c}{Model} & Type  & Design  & Search Time & Params {[}M{]}$\downarrow$  & MACs {[}G{]}$\downarrow$  & Epochs$\downarrow$  & Top-1 (\%)$\uparrow$  \\ 
\midrule

XCiT-N12~\cite{xcit}                           & Hybrid      & Manual &  -  & 3.0  & 0.5  & 400  & 69.9  \\
MobileViT-XS~\cite{mobilevit}                  & Hybrid      & Manual &  -  & 2.3  & 0.7  & 300  & 74.8  \\
EdgeViT-XXS~\cite{edgevit}                     & Hybrid      & Manual &  -  & 4.1  & 0.6  & 300  & 74.4  \\
MobileFormer-96M~\cite{mobileformer}           & Hybrid      & Manual &  -  & 4.6  & 0.1  & 450  & 72.8  \\
\cdashline{1-8}
EfficientFormerV2-S0~\cite{efficientFormerV2}  & Hybrid      & Auto   &  $>$ 8 GPU days  & 3.5  & 0.4  & 300  & $71.8^{\dag}$ \\ 
\rowcolor{Gray}
ESFormer-S0 (ours)                             & Hybrid      & Auto   &  0.75 GPU hours  & 3.5  & 0.4  & 300  & 75.5   \\
\midrule

DeiT-T~\cite{deit}                             & Transformer & Manual &  -  & 5.9  & 1.2  & 300  & 72.2  \\
XCiT-T12~\cite{xcit}                           & Hybrid      & Manual &  -  & 7.0  & 1.2  & 400  & 77.1  \\
MobileViT-S~\cite{mobilevit}                   & Hybrid      & Manual &  -  & 5.6  & 2.0  & 300  & 78.4  \\
EdgeViT-XS~\cite{edgevit}                      & Hybrid      & Manual &  -  & 6.7  & 1.1  & 300  & 77.5  \\
LeViT-128S~\cite{levit}                        & Hybrid      & Manual &  -  & 7.8  & 0.3  & 1000 & $76.6^{\diamond}$ \\
MobileFormer-151M~\cite{mobileformer}          & Hybrid      & Manual &  -  & 7.6  & 0.2  & 450  & 75.2  \\ 
Edge-NeXt-S~\cite{edgenext}                    & Hybrid      & Manual &  -  & 5.6  & 1.0  & 300  & 78.8  \\ 
\cdashline{1-8}
TF-TAS-Ti~\cite{dss}                           & Transformer & Auto   &  0.5 GPU days  & 5.9  & 1.4  & 300  & $75.3^{\diamond}$ \\
ViTAS-DeiT-A~\cite{vitas}                      & Transformer & Auto   &  $\sim$ 8 GPU days$\ddagger$ & 6.6  & 1.4  & 300  &  75.6 \\
GLiT-Tiny~\cite{glit}                          & Hybrid      & Auto   &  $>$ 10 GPU days$\ddagger$  & 7.2  & 1.4  & 1000  & $76.3^{\diamond}$ \\
BurgerFormer-Tiny~\cite{burger}                & Hybrid      & Auto   &  11 GPU days  &  10  &  1.0  &  300 &  78.0 \\
EfficientFormerV2-S1~\cite{efficientFormerV2}  & Hybrid      & Auto   &  $>$ 8 GPU days$\ddagger$  & 6.1  & 0.7  & 300  & $75.6^{\dag}$ \\
\rowcolor{Gray}
ESFormer-S1 (ours)                             & Hybrid      & Auto   &  0.75 GPU hours  & 5.9  & 0.9  & 300  & 78.8  \\
\midrule

LeViT-192~\cite{levit}                         & Hybrid      & Manual &  -  & 10.9 & 0.7  & 1000 & $80.0^{\diamond}$  \\
MobileFormer-508M~\cite{mobileformer}          & Hybrid      & Manual &  -  & 14.0 & 0.5  & 450  & 79.3  \\
XCiT-T24~\cite{xcit}                           & Hybrid      & Manual &  -  & 12.0 & 2.3  & 400  & 79.4  \\
EdgeViT-S~\cite{edgevit}                       & Hybrid      & Manual &  -  & 11.1 & 1.9  & 300  & 81.0  \\
\cdashline{1-8}
ViTAS-Twins-T~\cite{vitas}                     & Hybrid      & Auto   &  $\sim$ 8 GPU days$\ddagger$  & 13.8 & 1.4  & 300  & 79.4  \\
BurgerFormer-Small~\cite{burger}               & Hybrid      & Auto   &  11 GPU days  &  14.0  &  2.1  &  300  &  80.4 \\
EfficientFormerV2-S2~\cite{efficientFormerV2}  & Hybrid      & Auto   &  $>$ 8 GPU days$\ddagger$  & 12.6 & 1.3  & 300  & $78.31^{\dag}$ \\
\rowcolor{Gray}
ESFormer-S2 (ours)                             & Hybrid      & Auto   &  0.9 GPU hours  & 11.1 & 1.4  & 300  &  80.4    \\ 
\bottomrule
\end{tabular}
\caption{Results for ImageNet-1k. All models are tested with standard resolution 224x224 except for MobileViTs~\cite{mobilevit}, for which the resolution is 256x256. $\diamond$ Trained with distillation. $\dag$ Trained with original training configuration w/o distillation. $\ddagger$ Search time is a conservative estimate, actual values not reported in original papers. $\uparrow$ stands for the higher the better. $\downarrow$~stands for the lower the better.}
\label{tab:imagenet}
\end{table*}

To compare our approach with state-of-the-art architectures and NAS techniques, we conducted experiments using the proposed fine-grained search space described in Section~\ref{search_space}. The purpose of this search is to determine the optimal network architecture for different footprints, as outlined in~\cite{efficientFormerV2}. The three targeted model sizes were S0, S1, and S2, each with a maximum parameter count of 3.5, 6, and 12.5 millions respectively. The resulting family of architectures is named ESFormer, from the name of the proposed metric.

Table~\ref{tab:imagenet} shows the performance of state-of-the-art families of mobile architectures for different model sizes on ImageNet. The results in Table~\ref{tab:imagenet} are mainly taken from the original papers, with the exception of EfficientFormerV2 models which have been retrained with their original training configuration without distillation. \vspace{2pt}

\noindent \textbf{Comparison with hand-designed networks.} 
Our searched architectures prove to achieve higher accuracy with respect to the majority of hand-designed architectures. 

In particular, our S0 achieves a Top-1 accuracy of 75.5\%, outperforming all other architectures with similar or even slightly higher computational budgets. For medium-sized models, ESFormer-S1 performs on par with the best architecture, Edge-NeXt-S~\cite{edgenext}, with a Top-1 accuracy of 78.8\%. 

It is possible to notice that, for the largest computational budget, EdgeViT-S~\cite{edgevit} achieves slightly higher performance, but this comes at the expense of a 35\% increase in MACs with respect to our S2 network. Notably, for the medium-sized architectures where the computational budgets are comparable, we have exceeded the performance of EdgeViT-XS~\cite{edgevit} by more than 1\%. 

Overall, for similar parameters and MACs counts, we achieve the best Top-1 accuracy in all considered scenarios. \vspace{2pt}

\noindent \textbf{Comparison with NAS-designed networks.}
Comparing our algorithm with respect to other NAS methods, we can immediately appreciate the search speed of our approach. The search time for ESFormers is always less than 1 GPU hour, while most of the methods require several GPU days to design the final model. Remarkably, we decrease the search time by a factor 12x with respect to the previous fastest method, TF-TAS~\cite{dss}, while achieving more than 3\% increase in Top-1 accuracy with less MACs. 

BurgerFormer-Small~\cite{burger} is the only NAS-designed architecture able to obtain competitive results with respect to our S2 model in terms of accuracy. However, its search time is orders of magnitudes higher, and the network has significantly more parameters ($+26\%$) and MACs ($+50\%$). 

Notably, our architectures largely outperforms EfficientFormerV2, with an increase in Top-1 accuracy of more than 3\% across all model sizes.

\subsection{Ablation Study}

\noindent \textbf{Correlation in NAS Benchmarks.} The rank correlation between training-free metrics and the test accuracy on common benchmark datasets for NAS is an interesting aspect to consider. In particular, Table~\ref{tab:correlations} shows a comparison of the correlation between CIFAR-10 Top-1 accuracy and the rank given by different state-of-the-art training-free metrics on two common NAS benchmarks, NATS-Bench~\cite{nats} and NAS-Bench-101~\cite{bench-101}.

The NAS-Bench-101 dataset, which contains roughly 400,000 convolutional topologies, is used for comparison in Table~\ref{tab:correlations_topology}, while the NATS-Bench search space, containing over 30,000 architectures with same topology and different sizes, is used for the comparison in Table~\ref{tab:correlations_size}. 
\begin{table*}[t]
    \begin{subtable}{.49\linewidth}
     \centering
     \begin{tabular}{lcc} 
        \toprule
        \multicolumn{1}{c}{Metric} & \multicolumn{1}{c}{Kendall $\tau$ $\uparrow$} & \multicolumn{1}{c}{Spearman $\rho$ $\uparrow$} \\
        \midrule
        
        NASWOT~\cite{naswot} & \multicolumn{1}{c}{0.26} & \multicolumn{1}{c}{0.37} \\
    
        LogSynflow~\cite{freerea} & \multicolumn{1}{c}{0.31} & \multicolumn{1}{c}{0.45} \\
        
        \metricNameSpace (ours) & \multicolumn{1}{c}{\textbf{0.50}} & \multicolumn{1}{c}{\textbf{0.68}} \\
        \bottomrule
    \end{tabular}
    \caption{Correlation w.r.t. a \textit{topological} search space.}
    \label{tab:correlations_topology}
    \end{subtable}
    \begin{subtable}{.49\linewidth}
    \centering
    \begin{tabular}{lcc} 
        \toprule
        \multicolumn{1}{c}{Metric} & \multicolumn{1}{c}{Kendall $\tau$ $\uparrow$} & \multicolumn{1}{c}{Spearman $\rho$ $\uparrow$} \\
        \midrule
        
        NASWOT~\cite{naswot} & \multicolumn{1}{c}{0.45} & \multicolumn{1}{c}{0.63} \\
    
        LogSynflow~\cite{freerea} & \multicolumn{1}{c}{\textbf{0.76}} & \multicolumn{1}{c}{\textbf{0.92}} \\
        
        \metricNameSpace (ours) & \multicolumn{1}{c}{0.03} & \multicolumn{1}{c}{0.04} \\
        \bottomrule
    \end{tabular}
    \caption{Correlation w.r.t. a \textit{size} search space.}
    \label{tab:correlations_size}
    \end{subtable} 
\caption{Kendall and Spearman rank correlation between training-free metrics and CIFAR10 Top-1 (\%) accuracy, evaluated on a) NAS-Bench-101~\cite{bench-101} topological search space and b) NATS-Bench~\cite{nats} size search space. $\uparrow$ stands for the higher the better. \metricNameSpace shows to be particularly suitable for topology definition. On the other hand, it lacks the ability of dimensioning the architecture, where LogSynflow excels.}
\label{tab:correlations}
\end{table*}
\begin{table*}[t]
\setlength{\tabcolsep}{3.5pt}
\centering
\begin{tabular}{ccc|cc|c}
\toprule
LogSynflow & \metricNameSpace & Decoupling & Params {[}M{]}$\downarrow$ & MACs {[}G{]}$\downarrow$ & Top-1 (\%)$\uparrow$ \\
\midrule

\checkmark  & \xmark      &  \xmark      &  6.00  &  0.86  &  72.4   \\
\checkmark  & \checkmark  &  \xmark      &  5.92  &  0.94  &  75.7   \\
\checkmark  & \checkmark  &  \checkmark  &  5.97  &  0.96  &  \textbf{77.8}   \\
\bottomrule
\end{tabular}
\caption{Ablation on different configurations of the search algorithm with the extended search space containing Residual Bottlenecks. Top-1 (\%) accuracy on ImageNet-1k is reported. $\downarrow$ stands for the lower the better. $\uparrow$ stands for the higher the better.}
\label{tab:ablation}
\vspace{-.3cm}
\end{table*}

The results demonstrate the exceptional capability of \metricNameSpace in determining suitable network topologies, as shown by its high correlation with accuracy, which is almost two times greater than the one achieved by NASWOT~\cite{naswot}. Still, Table~\ref{tab:correlations_size} shows how \metricNameSpace lacks the ability to determine the size of the architecture, an area where LogSynflow~\cite{freerea} instead excels. Similar findings can be appreciated in Figure~\ref{fig:correlations}, which reports the CIFAR-10 Top-1 accuracy with respect to the rank given by \metricNameSpace (Figure~\ref{fig:entropy}) and by LogSynflow (Figure~\ref{fig:logsynflow}) in a topological and size search space respectively.

This also vouches for the complementarity of the two adopted metrics (see Figure~\ref{fig:correlations}).

\noindent \textbf{Search Ablation.} To better showcase the advantages of using \metricNameSpace as a search metric, we extend the search space with an additional topological choice by incorporating a standard Residual Bottleneck block~\cite{resnet}.

This block is not suitable for mobile-sized networks and it would be rightly overlooked by standard training-based NAS algorithms that rely solely on validation accuracy as a supervisory signal. However, training-free NAS approaches that employ proxy metrics could consistently choose this type of block due to lack of accuracy information, resulting in poor performances of the final architecture. Instead, we show that \metricNameSpace is able to overcome this limitation.

We ablate our decoupling algorithm by performing several searches with different combinations of metrics (see Table~\ref{tab:ablation}). In particular, we compare a search guided solely by LogSynflow, a search that seamlessly combines LogSynflow and \metricNameSpace and our proposed algorithm. The hardware constraints were set to a maximum of 6 millions parameters, focusing on medium sized candidates.

In Table~\ref{tab:ablation}, it can be observed that the straightforward combination of LogSynflow and \metricNameSpace as guiding metrics already results in a significantly higher accuracy compared to the baseline configuration that relies on LogSynflow only, with an improvement of over 3\%. The decoupled search strategy, allowing for specialisation of the metrics, leads to even higher-performing architectures, with an additional consistent improvement of 2\%.
\section{Limitations}
\label{limitations}

The limitations of the proposed approach should be acknowledged. Although the results demonstrate the efficacy of \metricNameSpace in discovering high-performing neural network topologies, it is a training-free metric and therefore only a proxy for the actual performance of the architecture. Hence, it is likely that, if larger computational resources are available, even better networks can be discovered by including training in the search process.

Additionally, while \metricNameSpace excels in identifying high-performing network topologies, it does not show ability in determining network dimensions (see Table~\ref{tab:correlations}) and must be combined with other metrics to obtain satisfying results.
\section{Conclusions}
\label{conclusions}

In this work, we present a novel efficient training-free NAS framework leveraging an original metric, \metricName, to guide the search process on a flexible and fine-grained search space.

\metricNameSpace demonstrates to be particularly suitable to design the topology of the networks and it is combined with LogSynflow to account for the architecture size in an original decoupled fashion. Decoupling the design of topology and size allows each metric to focus on its strengths, leading to a more targeted and precise search, and an overall higher accuracy of the searched models. 

The discovered family of tiny Hybrid Transformers, named ESFormers, proves to be competitive with respect to the state-of-the-art in neural network design. ESFormers outperform not only hand-designed networks but also training-based NAS approaches. Remarkably, the search time is reduced to less than 1 GPU hour, a 12x improvement with respect to the previous fastest NAS method.

Future research directions can involve the development of more precise proxies for the performance of the architectures and the extension of the training-free framework to more complex tasks such as Segmentation or Detection.
\section{Acknowledgements}
\label{acks}

This study was carried out within the FAIR - Future Artificial Intelligence Research and received funding from the European Union Next-GenerationEU (PIANO NAZIONALE DI RIPRESA E RESILIENZA (PNRR) – MISSIONE 4 COMPONENTE 2, INVESTIMENTO 1.3 – D.D. 1555 11/10/2022, PE00000013). This manuscript reflects only the authors’ views and opinions, neither the European Union nor the European Commission can be considered responsible for them.

{\small
\bibliographystyle{ieee_fullname}
\bibliography{egpaper_final}

\begin{thebibliography}{10}\itemsep=-1pt

\bibitem{zerocost}
Mohamed~S Abdelfattah, Abhinav Mehrotra, {\L}ukasz Dudziak, and Nicholas~Donald
  Lane.
\newblock Zero-cost proxies for lightweight nas.
\newblock In {\em ICLR}, 2020.

\bibitem{relu}
Abien~Fred Agarap.
\newblock Deep learning using rectified linear units (relu).
\newblock {\em arXiv preprint arXiv:1803.08375}, 2018.

\bibitem{xcit}
Alaaeldin Ali, Hugo Touvron, Mathilde Caron, Piotr Bojanowski, Matthijs Douze,
  Armand Joulin, Ivan Laptev, Natalia Neverova, Gabriel Synnaeve, Jakob
  Verbeek, et~al.
\newblock Xcit: Cross-covariance image transformers.
\newblock {\em Advances in neural information processing systems}, 2021.

\bibitem{freerea}
Niccol{\`o} Cavagnero, Luca Robbiano, Barbara Caputo, and Giuseppe Averta.
\newblock Freerea: Training-free evolution-based architecture search.
\newblock In {\em WACV}, 2023.

\bibitem{glit}
Boyu Chen, Peixia Li, Chuming Li, Baopu Li, Lei Bai, Chen Lin, Ming Sun, Junjie
  Yan, and Wanli Ouyang.
\newblock Glit: Neural architecture search for global and local image
  transformer.
\newblock In {\em Proceedings of the IEEE/CVF International Conference on
  Computer Vision}, pages 12--21, 2021.

\bibitem{tenas}
Wuyang Chen, Xinyu Gong, and Zhangyang Wang.
\newblock Neural architecture search on imagenet in four gpu hours: A
  theoretically inspired perspective.
\newblock In {\em ICLR}, 2021.

\bibitem{mobileformer}
Yinpeng Chen, Xiyang Dai, Dongdong Chen, Mengchen Liu, Xiaoyi Dong, Lu Yuan,
  and Zicheng Liu.
\newblock Mobile-former: Bridging mobilenet and transformer.
\newblock In {\em CVPR}, 2022.

\bibitem{edgevit}
Zekai Chen, Fangtian Zhong, Qi Luo, Xiao Zhang, and Yanwei Zheng.
\newblock Edgevit: Efficient visual modeling for edge computing.
\newblock In {\em WASA}, 2022.

\bibitem{randaugment}
Ekin~D Cubuk, Barret Zoph, Jonathon Shlens, and Quoc~V Le.
\newblock Randaugment: Practical automated data augmentation with a reduced
  search space.
\newblock In {\em CVPR}, 2020.

\bibitem{coatnet}
Zihang Dai, Hanxiao Liu, Quoc~V Le, and Mingxing Tan.
\newblock Coatnet: Marrying convolution and attention for all data sizes.
\newblock {\em NIPS}, 2021.

\bibitem{imagenet}
Jia Deng, Wei Dong, Richard Socher, Li-Jia Li, Kai Li, and Li Fei-Fei.
\newblock Imagenet: A large-scale hierarchical image database.
\newblock In {\em CVPR}, 2009.

\bibitem{nats}
Xuanyi Dong, Lu Liu, Katarzyna Musial, and Bogdan Gabrys.
\newblock Nats-bench: Benchmarking nas algorithms for architecture topology and
  size.
\newblock {\em IEEE transactions on pattern analysis and machine intelligence},
  2021.

\bibitem{gdas}
Xuanyi Dong and Yi Yang.
\newblock Searching for a robust neural architecture in four gpu hours.
\newblock In {\em CVPR}, 2019.

\bibitem{vit}
Alexey Dosovitskiy, Lucas Beyer, Alexander Kolesnikov, Dirk Weissenborn,
  Xiaohua Zhai, Thomas Unterthiner, Mostafa Dehghani, Matthias Minderer, Georg
  Heigold, Sylvain Gelly, et~al.
\newblock An image is worth 16x16 words: Transformers for image recognition at
  scale.
\newblock {\em arXiv preprint arXiv:2010.11929}, 2020.

\bibitem{nasvit}
Chengyue Gong, Dilin Wang, Meng Li, Xinlei Chen, Zhicheng Yan, Yuandong Tian,
  Vikas Chandra, et~al.
\newblock Nasvit: Neural architecture search for efficient vision transformers
  with gradient conflict aware supernet training.
\newblock In {\em International Conference on Learning Representations}, 2021.

\bibitem{levit}
Benjamin Graham, Alaaeldin El-Nouby, Hugo Touvron, Pierre Stock, Armand Joulin,
  Herv{\'e} J{\'e}gou, and Matthijs Douze.
\newblock Levit: a vision transformer in convnet's clothing for faster
  inference.
\newblock In {\em ICCV}, 2021.

\bibitem{pyramid}
Dongyoon Han, Jiwhan Kim, and Junmo Kim.
\newblock Deep pyramidal residual networks.
\newblock In {\em Proceedings of the IEEE conference on computer vision and
  pattern recognition}, pages 5927--5935, 2017.

\bibitem{resnet}
Kaiming He, Xiangyu Zhang, Shaoqing Ren, and Jian Sun.
\newblock Deep residual learning for image recognition.
\newblock In {\em Proceedings of the IEEE conference on computer vision and
  pattern recognition}, pages 770--778, 2016.

\bibitem{gelu}
Dan Hendrycks and Kevin Gimpel.
\newblock Gaussian error linear units (gelus).
\newblock {\em arXiv preprint arXiv:1606.08415}, 2016.

\bibitem{stochdepth}
Gao Huang, Yu Sun, Zhuang Liu, Daniel Sedra, and Kilian~Q Weinberger.
\newblock Deep networks with stochastic depth.
\newblock In {\em ECCV}, 2016.

\bibitem{ntk}
Arthur Jacot, Franck Gabriel, and Cl{\'e}ment Hongler.
\newblock Neural tangent kernel: Convergence and generalization in neural
  networks.
\newblock {\em Advances in neural information processing systems}, 2018.

\bibitem{snip}
Namhoon Lee, Thalaiyasingam Ajanthan, and Philip H.~S. Torr.
\newblock {SNIP:} single-shot network pruning based on connection sensitivity.
\newblock {\em CoRR}, 2018.

\bibitem{efficientFormerV2}
Yanyu Li, Ju Hu, Yang Wen, Georgios Evangelidis, Kamyar Salahi, Yanzhi Wang,
  Sergey Tulyakov, and Jian Ren.
\newblock Rethinking vision transformers for mobilenet size and speed.
\newblock {\em arXiv preprint arXiv:2212.08059}, 2022.

\bibitem{efficientformer}
Yanyu Li, Geng Yuan, Yang Wen, Eric Hu, Georgios Evangelidis, Sergey Tulyakov,
  Yanzhi Wang, and Jian Ren.
\newblock Efficientformer: Vision transformers at mobilenet speed.
\newblock {\em arXiv preprint arXiv:2206.01191}, 2022.

\bibitem{localvit}
Yawei Li, Kai Zhang, Jiezhang Cao, Radu Timofte, and Luc Van~Gool.
\newblock Localvit: Bringing locality to vision transformers.
\newblock {\em arXiv preprint arXiv:2104.05707}, 2021.

\bibitem{zenscore}
Ming Lin, Pichao Wang, Zhenhong Sun, Hesen Chen, Xiuyu Sun, Qi Qian, Hao Li,
  and Rong Jin.
\newblock Zen-nas: A zero-shot nas for high-performance image recognition.
\newblock In {\em ICCV}, 2021.

\bibitem{darts}
Hanxiao Liu, Karen Simonyan, and Yiming Yang.
\newblock Darts: Differentiable architecture search.
\newblock In {\em ICML}, 2018.

\bibitem{evo_nas_review}
Yuqiao Liu, Yanan Sun, Bing Xue, Mengjie Zhang, Gary~G Yen, and Kay~Chen Tan.
\newblock A survey on evolutionary neural architecture search.
\newblock {\em IEEE transactions on neural networks and learning systems},
  2021.

\bibitem{swin}
Ze Liu, Yutong Lin, Yue Cao, Han Hu, Yixuan Wei, Zheng Zhang, Stephen Lin, and
  Baining Guo.
\newblock Swin transformer: Hierarchical vision transformer using shifted
  windows.
\newblock In {\em ICCV}, 2021.

\bibitem{convnext}
Zhuang Liu, Hanzi Mao, Chao{-}Yuan Wu, Christoph Feichtenhofer, Trevor Darrell,
  and Saining Xie.
\newblock A convnet for the 2020s.
\newblock {\em CoRR}, 2022.

\bibitem{sgdr}
Ilya Loshchilov and Frank Hutter.
\newblock Sgdr: Stochastic gradient descent with warm restarts.
\newblock {\em arXiv preprint arXiv:1608.03983}, 2016.

\bibitem{adamw}
Ilya Loshchilov and Frank Hutter.
\newblock Decoupled weight decay regularization.
\newblock {\em arXiv preprint arXiv:1711.05101}, 2017.

\bibitem{edgenext}
Muhammad Maaz, Abdelrahman Shaker, Hisham Cholakkal, Salman Khan, Syed~Waqas
  Zamir, Rao~Muhammad Anwer, and Fahad Shahbaz~Khan.
\newblock Edgenext: efficiently amalgamated cnn-transformer architecture for
  mobile vision applications.
\newblock In {\em ECCV}, 2023.

\bibitem{mobilevit}
Sachin Mehta and Mohammad Rastegari.
\newblock Mobilevit: light-weight, general-purpose, and mobile-friendly vision
  transformer.
\newblock {\em arXiv preprint arXiv:2110.02178}, 2021.

\bibitem{naswot}
Joe Mellor, Jack Turner, Amos Storkey, and Elliot~J Crowley.
\newblock Neural architecture search without training.
\newblock In {\em ICML}, 2021.

\bibitem{grad_clip}
Razvan Pascanu, Tomas Mikolov, and Yoshua Bengio.
\newblock On the difficulty of training recurrent neural networks.
\newblock In {\em ICML}, 2013.

\bibitem{swish}
Prajit Ramachandran, Barret Zoph, and Quoc~V Le.
\newblock Searching for activation functions.
\newblock {\em arXiv preprint arXiv:1710.05941}, 2017.

\bibitem{rea}
Esteban Real, Alok Aggarwal, Yanping Huang, and Quoc~V Le.
\newblock Regularized evolution for image classifier architecture search.
\newblock In {\em Proceedings of the aaai conference on artificial
  intelligence}, 2019.

\bibitem{evo_large}
Esteban Real, Sherry Moore, Andrew Selle, Saurabh Saxena, Yutaka~Leon Suematsu,
  Jie Tan, Quoc~V Le, and Alexey Kurakin.
\newblock Large-scale evolution of image classifiers.
\newblock In {\em ICML}, 2017.

\bibitem{mobilenetV2}
Mark Sandler, Andrew Howard, Menglong Zhu, Andrey Zhmoginov, and Liang-Chieh
  Chen.
\newblock Mobilenetv2: Inverted residuals and linear bottlenecks.
\newblock In {\em Proceedings of the IEEE conference on computer vision and
  pattern recognition}, pages 4510--4520, 2018.

\bibitem{talking}
Noam Shazeer, Zhenzhong Lan, Youlong Cheng, Nan Ding, and Le Hou.
\newblock Talking-heads attention.
\newblock {\em arXiv preprint arXiv:2003.02436}, 2020.

\bibitem{inception}
Chenyang Si, Weihao Yu, Pan Zhou, Yichen Zhou, Xinchao Wang, and Shuicheng Yan.
\newblock Inception transformer.
\newblock {\em arXiv preprint arXiv:2205.12956}, 2022.

\bibitem{vitas}
Xiu Su, Shan You, Jiyang Xie, Mingkai Zheng, Fei Wang, Chen Qian, Changshui
  Zhang, Xiaogang Wang, and Chang Xu.
\newblock Vitas: Vision transformer architecture search.
\newblock In {\em ECCV}, 2022.

\bibitem{labelsmooth}
Christian Szegedy, Vincent Vanhoucke, Sergey Ioffe, Jon Shlens, and Zbigniew
  Wojna.
\newblock Rethinking the inception architecture for computer vision.
\newblock In {\em CVPR}, 2016.

\bibitem{efficientnet}
Mingxing Tan and Quoc Le.
\newblock Efficientnet: Rethinking model scaling for convolutional neural
  networks.
\newblock In {\em ICML}, 2019.

\bibitem{synflow}
Hidenori Tanaka, Daniel Kunin, Daniel~L Yamins, and Surya Ganguli.
\newblock Pruning neural networks without any data by iteratively conserving
  synaptic flow.
\newblock {\em NIPS}, 2020.

\bibitem{deit}
Hugo Touvron, Matthieu Cord, Matthijs Douze, Francisco Massa, Alexandre
  Sablayrolles, and Herv{\'e} J{\'e}gou.
\newblock Training data-efficient image transformers \& distillation through
  attention.
\newblock In {\em ICML}, 2021.

\bibitem{fisher}
Jack Turner, Elliot~J. Crowley, Gavin Gray, Amos~J. Storkey, and Michael F.~P.
  O'Boyle.
\newblock Blockswap: Fisher-guided block substitution for network compression.
\newblock {\em CoRR}, 2019.

\bibitem{attention}
Ashish Vaswani, Noam Shazeer, Niki Parmar, Jakob Uszkoreit, Llion Jones,
  Aidan~N. Gomez, Lukasz Kaiser, and Illia Polosukhin.
\newblock Attention is all you need.
\newblock {\em NIPS}, 2017.

\bibitem{grasp}
Chaoqi Wang, Guodong Zhang, and Roger~Baker Grosse.
\newblock Picking winning tickets before training by preserving gradient flow.
\newblock {\em CoRR}, 2020.

\bibitem{resnext}
Saining Xie, Ross Girshick, Piotr Doll{\'a}r, Zhuowen Tu, and Kaiming He.
\newblock Aggregated residual transformations for deep neural networks.
\newblock In {\em CVPR}, 2017.

\bibitem{burger}
Longxing Yang, Yu Hu, Shun Lu, Zihao Sun, Jilin Mei, Yinhe Han, and Xiaowei Li.
\newblock Searching for burgerformer with micro-meso-macro space design.
\newblock In {\em ICML}, 2022.

\bibitem{bench-101}
Chris Ying, Aaron Klein, Eric Christiansen, Esteban Real, Kevin Murphy, and
  Frank Hutter.
\newblock Nas-bench-101: Towards reproducible neural architecture search.
\newblock In {\em ICML}, 2019.

\bibitem{cutmix}
Sangdoo Yun, Dongyoon Han, Seong~Joon Oh, Sanghyuk Chun, Junsuk Choe, and
  Youngjoon Yoo.
\newblock Cutmix: Regularization strategy to train strong classifiers with
  localizable features.
\newblock In {\em ICCV}, 2019.

\bibitem{mixup}
Hongyi Zhang, Moustapha Cisse, Yann~N Dauphin, and David Lopez-Paz.
\newblock mixup: Beyond empirical risk minimization.
\newblock {\em arXiv preprint arXiv:1710.09412}, 2017.

\bibitem{rand_erasing}
Zhun Zhong, Liang Zheng, Guoliang Kang, Shaozi Li, and Yi Yang.
\newblock Random erasing data augmentation.
\newblock In {\em Proceedings of the AAAI conference on artificial
  intelligence}, 2020.

\bibitem{dss}
Qinqin Zhou, Kekai Sheng, Xiawu Zheng, Ke Li, Xing Sun, Yonghong Tian, Jie
  Chen, and Rongrong Ji.
\newblock Training-free transformer architecture search.
\newblock In {\em Proceedings of the IEEE/CVF Conference on Computer Vision and
  Pattern Recognition}, pages 10894--10903, 2022.

\bibitem{reinforcement_nas}
Barret Zoph and Quoc~V. Le.
\newblock Neural architecture search with reinforcement learning.
\newblock {\em CoRR}, 2016.

\end{thebibliography}
}

\end{document}